%% file: main.tex
\documentclass{article}

\usepackage[preprint]{icml2026} 

\input{preamble}

\begin{document}

\input{01_title}

\input{02_abstract}
\input{03_introduction}
\input{04_problem}
\input{05_methods}

\input{06_evidence}
\input{07_actions}
\input{08_views}
\input{09_conclusions}


\nocite{langley00}


\bibliography{reference}
\bibliographystyle{icml2026}

\input{10_appendix}

\end{document}

%% file: preamble.tex
\usepackage{microtype}
\usepackage{graphicx}
\usepackage{subcaption}
\usepackage{booktabs} 

\definecolor{ICMLBlue}{RGB}{25, 88, 166}
\definecolor{ICMLPurple}{RGB}{150, 0, 80}
\usepackage[colorlinks,linkcolor=ICMLPurple,citecolor=ICMLBlue,urlcolor=black]{hyperref}

\usepackage{amsmath}
\usepackage{amssymb}
\usepackage{mathtools}
\usepackage{amsthm}
\usepackage{multirow}
\usepackage{xspace}
\usepackage{enumitem}
\usepackage{array}
\usepackage{xurl}
\usepackage{bm}
\usepackage{array}

\usepackage[capitalize,noabbrev]{cleveref}

\theoremstyle{plain}

\theoremstyle{definition}

\theoremstyle{remark}

\usepackage[textsize=tiny]{todonotes}

\usepackage{tikz}
\usetikzlibrary{mindmap,trees,shadows}
\usepackage[edges]{forest}

\definecolor{academicblue}{RGB}{54, 95, 145}

\definecolor{c1}{HTML}{fbdedc} 
\definecolor{c2}{HTML}{C5E0B4} 
\definecolor{c3}{HTML}{FBF2CE} 
\definecolor{c4}{HTML}{cf3e3e} 
\definecolor{c5}{HTML}{67a550} 
\definecolor{c6}{HTML}{bf9000} 
\definecolor{c7}{HTML}{bec4d0} 

\tikzset{forked edges/.style={edge from parent forked}}

\renewcommand{\paragraph}[1]{\noindent\textbf{#1}~}

\usepackage{xcolor}
\usepackage{tcolorbox}

\definecolor{promptpurple}{HTML}{7A1E8A}
\definecolor{promptblue}{HTML}{7EC3EA}

\newcounter{promptcount}
\renewcommand{\thepromptcount}{\arabic{promptcount}}

\newtcolorbox{promptbox}[2][]{%
  colframe=promptblue,
  colback=white,
  colbacktitle=promptblue,
  coltitle=white,
  title=\textbf{Prompt \refstepcounter{promptcount}\thepromptcount: #2},
  arc=4pt,
  boxrule=1pt,
  left=6pt,
  right=6pt,
  top=6pt,
  bottom=6pt,
  fonttitle=\bfseries,
  label=#1
}

%% file: 01_title.tex
\icmltitlerunning{LLMs Should Incorporate Explicit Mechanisms for Human Empathy}

\twocolumn[
  \icmltitle{LLMs Should Incorporate Explicit Mechanisms for Human Empathy}



  

  \begin{icmlauthorlist}
    \icmlauthor{Xiaoxing You}{HDU}
    \icmlauthor{Qiang Huang}{HITsz}
    \icmlauthor{Jun Yu}{HITsz}
  \end{icmlauthorlist}

  \icmlaffiliation{HDU}{School of Computer Science, Hangzhou Dianzi University}
  \icmlaffiliation{HITsz}{School of Intelligence Science and Engineering, Harbin Institute of Technology (Shenzhen)}

  \icmlcorrespondingauthor{Qiang Huang}{huangqiang@hit.edu.cn}
  \icmlcorrespondingauthor{Jun Yu}{yujun@hit.edu.cn}

  \icmlkeywords{Machine Learning, ICML}

  \vskip 0.3in
]



\printAffiliationsAndNotice{}  

%% file: 02_abstract.tex
\begin{abstract}
This paper argues that Large Language Models (LLMs) should incorporate explicit mechanisms for human empathy.
As LLMs become increasingly deployed in high-stakes human-centered settings, their success depends not only on correctness or fluency but on faithful preservation of human perspectives. 
Yet, current LLMs systematically fail at this requirement: even when well-aligned and policy-compliant, they often attenuate affect, misrepresent contextual salience, and rigidify relational stance in ways that distort meaning.
We formalize empathy as an observable behavioral property: the capacity to model and respond to human perspectives while preserving intention, affect, and context. 
Under this framing, we identify four recurring mechanisms of empathic failure in contemporary LLMs--sentiment attenuation, empathic granularity mismatch, conflict avoidance, and linguistic distancing--arising as structural consequences of prevailing training and alignment practices.
We further organize these failures along three dimensions: cognitive, cultural, and relational empathy, to explain their manifestation across tasks.
Empirical analyses show that strong benchmark performance can mask systematic empathic distortions, motivating empathy-aware objectives, benchmarks, and training signals as first-class components of LLM development.
\end{abstract}


%% file: 03_introduction.tex
\section{Introduction}
\label{sec:intro}

\begin{figure}
  \centering
  \includegraphics[width=0.99\linewidth]{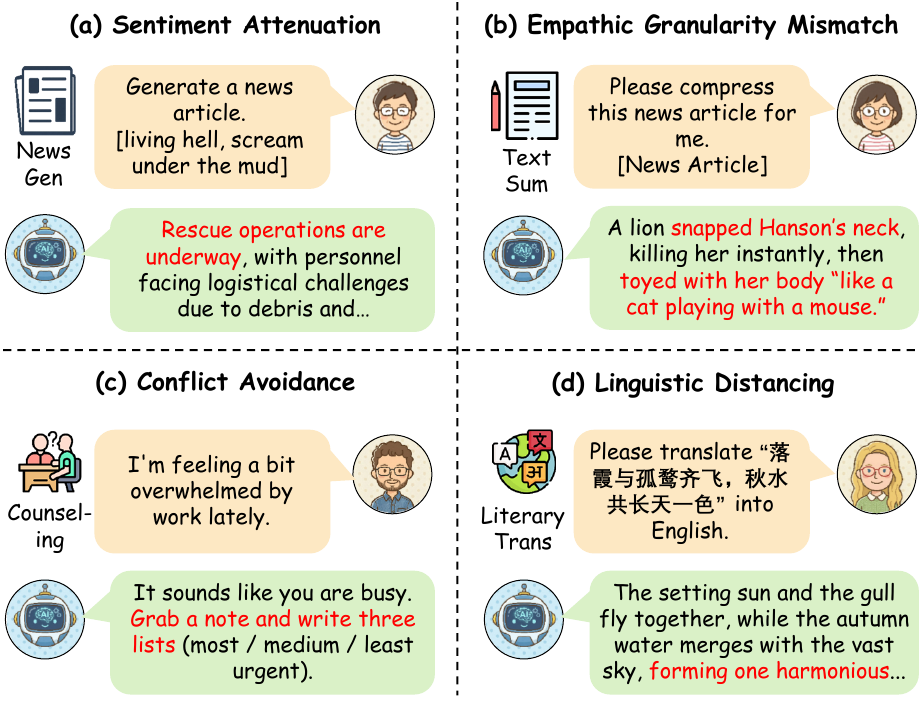}
  \caption{\textbf{Four recurring mechanisms underlying empathic failure in contemporary LLM outputs.} (a) Sentiment Attenuation: the model replaces emotionally charged language with neutral, official wording; (b) Empathic Granularity Mismatch: the model retains graphic detail where calibration is needed; (c) Conflict Avoidance: the model bypasses emotional tension and jumps directly to task-oriented advice; (d) Linguistic Distancing: the translation flattens the poem's imagery into plain prose.}
  \label{fig:mechanisms}
  \vspace{-1.0em}
\end{figure}

Large Language Models (LLMs) are no longer experimental assistants. 
They are deployed at scale as interfaces for a wide range of applications, including news production \cite{tohidi2025rethinking, liu2025ai, xu2025ctrlnews, odabacsi2025unraveling, sinha2025qa, sun2024diversinews, sun2025prism, sahu2025guide}, education \cite{kim2025histochat, breen2025large, park2025ask}, and psychological support \cite{de2025introducing, wang2025chatthero, xue2026mindchat}.
In these settings, correctness and fluency are necessary but insufficient. 
Increasingly, success depends on whether model outputs faithfully represent \emph{human perspectives}--including what users intend, feel, and consider salient in context \cite{sorin2024large, lee2024large, welivita2024large, sun2025text, sun2025one}. 

We argue that LLMs should incorporate explicit mechanisms for human empathy.
Empathy, as we define it, is not an internal mental state or stylistic embellishment, but an \emph{observable behavioral property}: the capacity to model and respond to human perspectives while preserving intention, affect, and contextual requirements in generated outputs.
When this capacity is absent, models may remain helpful, safe, and policy-aligned, yet still distort meaning in ways that materially affect real-world outcomes.

This limitation is not incidental.
Modern LLMs \cite{openai2025gpt52, anthropic2025claudeSonnet45, google2025gemini3, liu2025deepseekv32, bai2025qwen3vl, hao2026unix} are trained through large-scale pretraining for broad competence, followed by post-training alignment via instruction tuning~\cite{wei2022finetuned, ouyang2022training, feng2025don} and preference optimization~\cite{schulman2017ppo, rafailov2023dpo, shao2024grpo, hong2024orpo, zhou2024holistic} to improve helpfulness, safety, and risk mitigation.
While effective at producing fluent and compliant outputs, these objectives systematically favor calmness and neutrality, often suppressing necessary affect, perspective specificity, and relational engagement. 
Models may therefore satisfy alignment criteria while misrepresenting human perspectives where fidelity matters most.

To make these failure modes explicit, we summarize four recurring mechanisms through which current LLMs exhibit \emph{deficient empathy}. Specifically, we identify: 
\begin{itemize}[nosep, left=0pt]
  \item \textbf{Sentiment Attenuation}, where emotional intensity or urgency is systematically downscaled \cite{bardol2025chatgpt}; 
  
  \item \textbf{Empathic Granularity Mismatch}, where the level of affective or experiential detail is miscalibrated \cite{manevich2023multi};
  
  \item \textbf{Conflict Avoidance}, where legitimate tensions are deflected in favor of superficial harmony \cite{sharma2024towards}; 
  
  \item \textbf{Linguistic Distancing}, where depersonalized framing increases interpersonal distance \cite{nook2017linguistic}.
\end{itemize}
As illustrated in Figure~\ref{fig:mechanisms}, these mechanisms are not edge cases: they consistently emerge under current training and alignment regimes across application domains~\cite{kirk2024understanding, xu2025consistency}.

To explain how these mechanisms manifest differently across tasks, we characterize empathy along three dimensions: \textbf{cognitive}, \textbf{cultural}, and \textbf{relational empathy}. 
Cognitive empathy governs perspective-taking and the preservation of informational salience~\cite{cuff2016empathy, healey2018cognitive, reniers2022cognitive}; cultural empathy captures sensitivity to norms, symbols, and aesthetic context~\cite{cuff2016empathy, eichbaum2023empathy}; and relational empathy supports trust, cooperation, and constructive engagement over interaction~\cite{eisenberg2010empathy, cuff2016empathy, weisz2018motivated, van2020towards}.
This decomposition clarifies why empathic failures vary across tasks while sharing common underlying causes.

We substantiate this position through empirical analyses on representative high-stakes tasks, including news generation, literary translation, and counseling-style response generation \cite{grusky2018newsroom, wang2023findings, wang2023guofeng}.
Across all settings, frontier LLMs exhibit consistent empathic distortions despite strong performance on conventional metrics. 
These failures reveal alignment trade-offs that remain invisible under existing evaluation frameworks.

Based on these findings, we outline concrete directions for integrating empathic fidelity into LLM development pipelines. 
We advocate for empathy-aware alignment objectives, task-grounded benchmarks that expose mechanism-level failures, and targeted data curation that supervises perspective preservation, affective calibration, and relational stance.
Taken together, this position reframes empathy from a stylistic byproduct into a measurable and optimizable property of aligned language models.

%% file: 04_problem.tex
\section{Problem Formulation}
\label{sec:problem}

\subsection{Defining Empathy for LLMs}
\label{sec:problem:define}

In psychology, empathy broadly refers to perspective-taking and context-sensitive responding in social interaction \cite{singer2009social, zaki2014empathy, levett2019systematic}. 
For LLMs, we define empathy as an \emph{observable behavioral property}: the capacity to model and respond to human perspectives while preserving intention, affect, and contextual requirements in generated outputs. 

This framing treats empathy as a form of representational fidelity in language generation. 
Beyond factual correctness and task completion, empathetic outputs should preserve the \emph{salience structure} and \emph{pragmatic implications} that matter to users, while appropriately calibrating affective granularity and relational framing to the interaction context~\cite{schulman2017ppo, rafailov2023dpo, kopf2023openassistant, shao2024grpo, hong2024orpo}. 
Such fidelity is especially critical in high-stakes applications, including mental health support~\cite{lozoya2025leveraging, xue2026mindchat}, news summarization~\cite{you2025event, odabacsi2025unraveling}, and workplace communication~\cite{navarro2025email, miura2025understanding}, where outputs may be factually correct yet fail to convey urgency, misrepresent perspectives, or adopt inappropriate interpersonal distance.

\subsection{Four Mechanisms of Empathic Failure}
\label{sec:problem:mech}

Despite producing fluent and policy-compliant outputs, LLMs often fail to capture what humans \textit{mean}, \textit{feel}, and \textit{need} in context~\cite{sorin2024large, chen2024emotionqueen, he2025plan, kim2025being}. 
We identify four recurring mechanisms underlying such empathic deficits.

\input{tables/metrics}

\paragraph{Notation}
Let $n$ denote the number of evaluated instances. 
For each instance $i \in \{1,\cdots,n\}$, let $\bm{t}^{(i)}$ denote the reference text and $\bm{g}^{(i)}$ the model-generated output.

\paragraph{Sentiment Attenuation (SA)} 
Sentiment Attenuation describes the systematic reduction of emotional intensity, urgency, or moral weight in model outputs, even when the underlying facts are preserved.
It commonly arises when LLMs reframe human experiences into ``neutral'' summaries that appear calm, balanced, and professionally flattened \cite{bardol2025chatgpt, parmar2025emotionally}.
This pattern likely stems from alignment procedures that implicitly penalize strong affective language \cite{kirk2024understanding}, suppressing emotional intensity even when it is semantically necessary. 

We quantify this effect using the \emph{sentiment attenuation rate}:
\vspace{-0.25em}
\begin{equation}
\label{eq:sa}
  SA = \frac{1}{n} \sum_{i=1}^{n} \frac{E(\bm{t}^{(i)}) - E(\bm{g}^{(i)})}{E(t^{(i)})}
  \vspace{-0.25em}
\end{equation}
where $E(\cdot) \in [0, 1]$ denotes an affect intensity scorer; 0 indicates emotionally flat and affectively neutral language, and 1 indicates highly intense and vividly expressed emotion.

\paragraph{Empathic Granularity Mismatch (EGM)} 
Empathic Granularity Mismatch captures miscalibration in the level of affective or experiential detail expressed in outputs, particularly in sensitive or distressing contexts \cite{manevich2023multi}.
Models may produce overly vague abstractions or unnecessarily explicit descriptions~\cite{peters2025generalization}, a failure especially visible in news summarization involving violence or trauma~\cite{zhang2023summit, sahu2025guide}.
This miscalibration arises because pretraining emphasizes language modeling and commonsense acquisition \cite{wei2022finetuned, porada2022does}, while post-training targets alignment and reasoning \cite{shao2024grpo, hong2024orpo}, leaving descriptive granularity underspecified.

We measure empathic granularity mismatch via a \emph{granularity mismatch score}, which is defined as:
\vspace{-0.25em}
\begin{equation}
\label{eq:GM}
  EGM=\frac{1}{n} \sum_{i=1}^{n} {\Big|\frac{a^{(i)} - b^{(i)}}{a^{(i)}+b^{(i)}} \Big|},
  \vspace{-0.25em}
\end{equation}
where $a^{(i)}=\tfrac{U(\bm{t}^{(i)})}{d^{(i)}}$ and $b^{(i)}=\tfrac{U(\bm{g}^{(i)})}{q^{(i)}}$ denote the densities of disturbing or overly explicit content in the reference text $\bm{t}^{(i)}$ and model output $\bm{g}^{(i)}$, respectively; $U(\cdot)$ is a discomfort counter; and $d^{(i)}$ and $q^{(i)}$ are sentence counts.

\paragraph{Conflict Avoidance (CA)} 
Conflict Avoidance refers to the tendency to minimize, obscure, or prematurely resolve interpersonal tension rather than engaging with it directly~\cite{sharma2024towards, cheng2025elephant}.
A typical manifestation is premature solution-giving, where models bypass clarification of competing perspectives and trade-offs. This behavior is likely reinforced by alignment objectives that reward agreeableness and controversy reduction~\cite{rottger2024xstest}.

We measure this tendency via the \emph{conflict avoidance rate}, which is the proportion of instances where a model prematurely resolves tension through solution-giving or harmony-seeking, rather than engaging in clarification or perspective-specific inquiry. Formally,
\vspace{-0.25em}
\begin{equation}
\label{eq:ca}
  CA=\frac{1}{n} \sum_{i=1}^{n} s^{(i)},
  \vspace{-0.25em}
\end{equation}
where $s^{(i)} = J(\bm{g}^{(i)}) \in \{0,1\}$ is a binary indicator produced by a conflict-avoidance discriminator $J(\cdot)$, with $s^{(i)}=1$ denoting conflict avoidance and $s^{(i)}=0$ denoting faithful engagement with the underlying tension.

\paragraph{Linguistic Distancing (LD)} 
Linguistic Distancing refers to the use of abstract, indirect, or depersonalized language, such as passive constructions, nominalizations, and hedging, that reduces emotional immediacy \cite{danescu2013computational, nook2017linguistic, minnema2022dead}.
This manifests as a shift from agent-centered narratives to procedural descriptions that obscure agency and accountability, likely driven by alignment objectives favoring neutral, professional registers that minimize affective or evaluative directness \cite{liu2023trustworthy, huang2024collective}.

We define the \emph{linguistic distancing score} as:
\vspace{-0.25em}
\begin{equation} \label{eq:LD}
  LD=\frac{1}{n} \sum_{i=1}^{n} \Big\| \frac{\bm{d}_{t}^{(i)} - \bm{\mu}_t}{\bm{\sigma}_t} - \frac{\bm{d}_{g}^{(i)} - \bm{\mu}_g}{\bm{\sigma}_g} \Big\|_{2},
  \vspace{-0.25em}
\end{equation}
where $\bm{d}_{t}^{(i)}, \bm{d}_{g}^{(i)} \in \mathbb{R}^{3}$ are distancing-density vectors for the reference text $\bm{t}^{(i)}$ and model output $\bm{g}^{(i)}$, computed along nominalization, impersonal subject framing, and agentless constructions; and $(\bm{\mu}_t,\bm{\sigma}_t)$ and $(\bm{\mu}_g,\bm{\sigma}_g)$ are the corresponding means and standard deviations used for standardization.

\paragraph{Remarks}
These four mechanisms provide a structured lens for analyzing empathic failures in current LLMs. They are not mutually exclusive and may co-occur within a single response.
Table~\ref{tab:metrics} summarizes the range and interpretation of each metric.
Explicitly identifying them enables targeted evaluation and informs principled interventions to improve empathic fidelity in aligned LLM systems.

%% file: tables/metrics.tex
\begin{table*}[t]
\centering
\small
\renewcommand{\arraystretch}{1.3}
\caption{Definitions, ranges, and interpretations of the four empathic failure metrics.}
\label{tab:metrics}
\resizebox{0.99\textwidth}{!}{
\begin{tabular}{lll}
  \toprule
  \textbf{Metric} & \textbf{Range} & \textbf{Interpretation} \\
  \midrule
  \textbf{Sentiment Attenuation (SA)} & $(-\infty,1]$ & 0: no attenuation; 1: strong affect flattening; negative: more intense than reference \\
  \textbf{Empathic Granularity Mismatch (EGM)} & $[0,1]$ & 0: close alignment in detail density; 1: severe mismatch \\
  \textbf{Conflict Avoidance (CA)} & $[0,1]$ & 0: no conflict avoidance; 1: all responses avoid conflict \\
  \textbf{Linguistic Distancing (LD)} & $[0,+\infty)$ & 0: no distancing difference; larger: more abstract and depersonalized \\
  \bottomrule
\end{tabular}}
\end{table*}

%% file: 05_methods.tex
\section{Empathy Dimensions}
\label{sec:method}

\input{figures/taxonomy}

Having defined empathy and identified four failure mechanisms in Section~\ref{sec:problem}, we now characterize how empathic deficits affect real-world LLM deployments~\cite{sorin2024large}.
We organize human-facing requirements along three complementary dimensions: \textbf{cognitive}, \textbf{cultural}, and \textbf{relational empathy}, which govern factual salience, contextual meaning, and interaction dynamics, respectively.
In this section, we define each dimension and illustrate it through representative task examples (Figure~\ref{fig:empathy_taxonomy_full}).

\subsection{Cognitive Empathy}
\label{sec:method:cognitive}

Cognitive empathy denotes an LLM's ability to model a human perspective, including beliefs, intentions, and knowledge state, and to preserve what is salient and meaningful from that perspective in generated outputs \cite{reniers2022cognitive}.
In human-facing tasks, faithful generations require not only factual correctness, but also preservation of contextual salience and implications as recognized by users \cite{hu2025learning, kim2025histochat, shi2025social, liu2025ai, wang2025generating}.

Without cognitive empathy, models may produce factually accurate outputs that nonetheless misrepresent urgency, distress, or priority, thereby distorting user-relevant meaning.
This dimension is therefore critical in tasks where language mediates public understanding or resource allocation. Representative tasks include:
\begin{itemize}[nosep, left=0pt]
  \item \textbf{Text Summarization:} Accurate summaries may still suppress urgency or affective significance if salience is misrepresented \cite{zhang2024benchmarking, xu2025evaluating, yuan2025understanding, odabacsi2025unraveling, sinha2025qa}. 
  
  \item \textbf{History Education:} Modeling historical actors' motives and constraints prevents presentist reinterpretation \cite{lim2025rethinking, bungum2025if, du2025role, santamaria2025critical, breen2025large}. 
  
  \item \textbf{Welfare Assessment:} Eligibility judgments require preserving claimants' contextual constraints beyond generic criteria \cite{dong2025heterogeneous, coz2025policymaking, shi2025social, karr2025behind, kong2026generative}.
  
  \item \textbf{News Generation:} Reporting sensitive events demands calibrated detail to preserve human impact without distortion \cite{diakopoulos2024generative, tohidi2025rethinking, liu2025ai, tseng2025ownership, xu2025ctrlnews}.
\end{itemize}

Across these tasks, failures in cognitive empathy commonly manifest as \textbf{sentiment attenuation}, \textbf{empathic granularity mismatch}, and \textbf{linguistic distancing}, yielding outputs that are correct yet misleading in salience.

\subsection{Cultural Empathy}
\label{sec:method:cultural}

Cultural empathy captures an LLM's ability to attend to norms, cultural symbols, and contextual conventions, preserving cultural meaning rather than defaulting to homogenized expression~\cite{eichbaum2023empathy}.
In creative and interpretive tasks, output quality depends not only on semantic accuracy but also on stylistic nuance, implicit values, and cultural specificity \cite{zhang2025good, gurung2025learning, sands2025evaluation}.

Without cultural empathy, outputs become stylistically flat, replacing distinctive voices with generic phrasing that erases cultural identity and artistic intent.
This dimension is central in tasks where language mediates aesthetic experience and cultural identity. Representative tasks include:
\begin{itemize}[nosep, left=0pt]
  \item \textbf{Literary Translation:} Preserving register, metaphor, and culturally grounded connotations \cite{macken2024machine, wu2025perhaps, zhang2025litransproqa, zhang2025good}.
  
  \item \textbf{Literary Generation:} Maintaining coherent style and avoiding generic narrative templates \cite{feng2025ssgen, Teleki2025survey, xia2025storywriter, yang2025storyllava, gurung2025learning, wang2025generating}.
  
  \item \textbf{Art Criticism:} Interpreting symbolism and historical context beyond template-based evaluation \cite{bin2024gallerygpt, qi2025photographer, banting2025simulated, khadangi2025cognartive, arita2025assesing, sands2025evaluation}.
\end{itemize}

Cultural empathy failures primarily arise through \textbf{empathic granularity mismatch} and \textbf{linguistic distancing}, producing outputs that are fluent but culturally impoverished.

\subsection{Relational Empathy}
\label{sec:method:relational}

Relational empathy concerns an LLM’s ability to adapt responses to interaction partners and maintain constructive relational dynamics over time \cite{van2020towards}.
Unlike cognitive and cultural empathy, which focus on representation, relational empathy emphasizes interactional calibration, supporting trust, cooperation, and psychological safety~\cite{main2017interpersonal, eichbaum2025re}.

Without relational empathy, interactions become rigid or misaligned with users' social and emotional needs, undermining engagement and task effectiveness.
This dimension is essential in collaborative settings where outcomes depend on sustained user engagement and psychological safety. 
Representative tasks include:
\begin{itemize}[nosep, left=0pt]
  \item \textbf{Psychological Counseling:} Balancing affective validation with appropriate boundaries \cite{sharma2021towards, wang2025can, lozoya2025leveraging, zeng2025psychcounsel, zhao2025zhixin, de2025introducing, xue2026mindchat}.
  
  \item \textbf{Workplace Communication:} Navigating hierarchy, face-saving, and professional norms without excessive softening or deflection \cite{lu2024corporate, miura2025understanding, navarro2025email, li2025emails, hu2025your, singh2025inclusive}.
  
  \item \textbf{Conflict Mediation:} Engaging disagreement constructively without premature harmony-seeking \cite{govers2024aidriven, tessler2024ai, chen2025simulating, juvino2025can, sakurai2025exploring, li2025moderation}.
\end{itemize}

Failures in relational empathy typically manifest as \textbf{conflict avoidance} and \textbf{linguistic distancing}, yielding responses that appear polite yet undermine trust and collaboration.

\input{tables/Classification_table}

\paragraph{Remarks}
These three dimensions characterize empathic behavior in LLMs: cognitive empathy preserves user-relevant salience, cultural empathy preserves contextual meaning, and relational empathy preserves cooperative dynamics.
Table~\ref{tab:dim-scena-mapping} maps representative tasks to empathic failure mechanisms, guiding the task-grounded evaluations that follow.

%% file: figures/taxonomy.tex
\tikzstyle{my-box}=[
    rectangle,
    draw=gray!50,
    rounded corners,
    text opacity=1,
    minimum height=2em,
    minimum width=5em,
    inner sep=5pt,
    inner ysep=8pt,
    align=left,
    fill opacity=0.30,
    line width=0.5pt,
]

\tikzstyle{leaf}=[
    my-box,
    minimum height=2em,
    fill=gray!5,
    text=black,
    align=left, 
    font=\normalsize,
    inner xsep=5pt,
    inner ysep=6pt,
    line width=1.0pt,
]

\tikzset{forked edges/.style={edge from parent forked}}

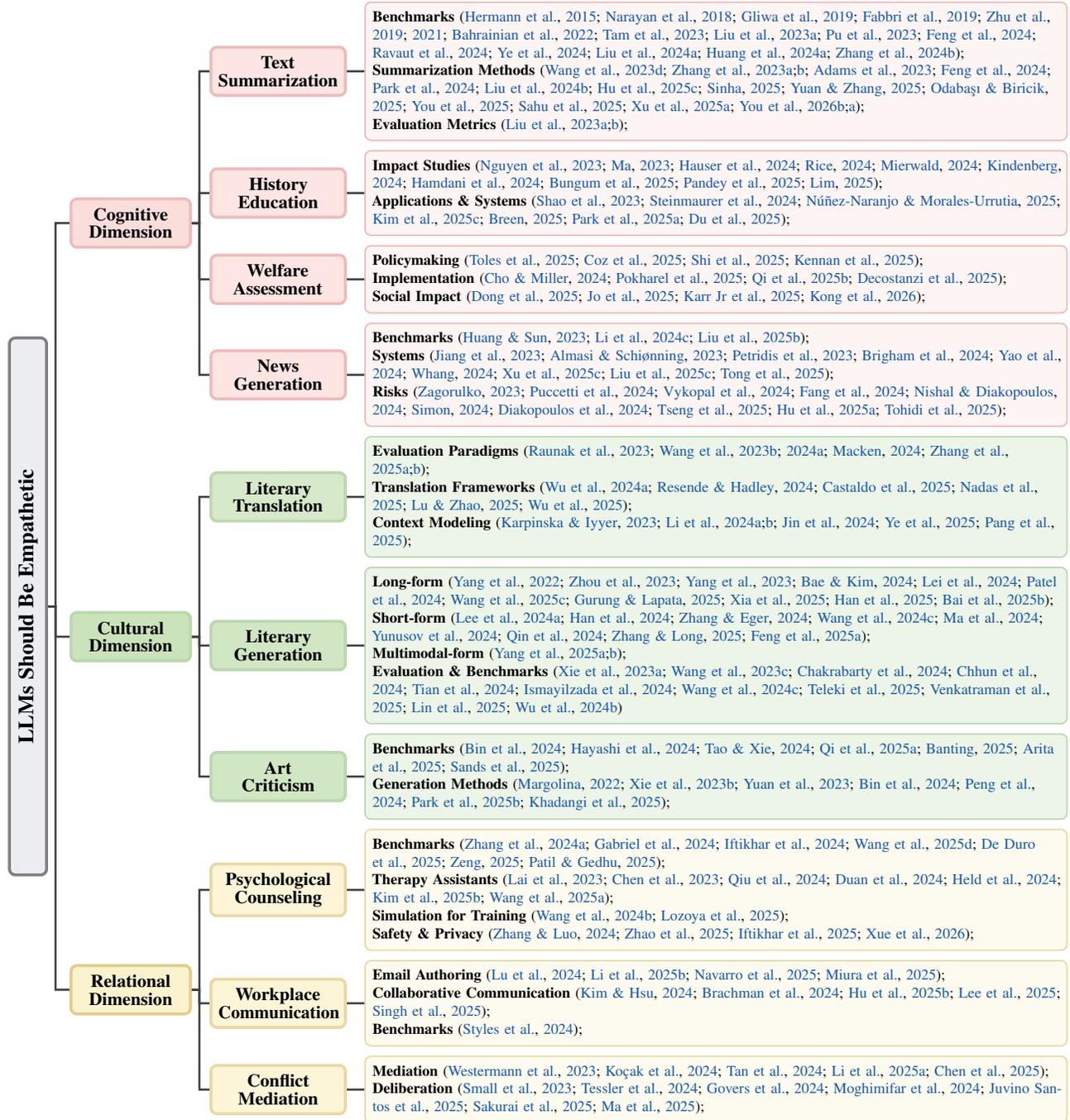
\begin{figure*}[!t]
  \centering
  \resizebox{\textwidth}{!}{
    \begin{forest}
      forked edges,
      for tree={
        grow=east,
        reversed=true,
        anchor=west,          
        parent anchor=east,   
        child anchor=west,    
        base=center,
        font=\large,
        rectangle,
        draw=gray,
        rounded corners,
        text centered, 
        minimum width=8em,
        edge+={darkgray, line width=0.5mm},
        s sep=6pt,
        inner xsep=5pt,
        inner ysep=6pt,
        line width=1.0pt,
        text width=35em, 
        ver/.style={rotate=90, child anchor=north, parent anchor=south, anchor=center},
      },
      where level=1{font=\normalsize,text width=7em}{},
      where level=2{font=\normalsize,text width=8em}{},
      where level=3{font=\normalsize,text width=46em}{},
      [
        {\Large \textbf{LLMs Should Be Empathetic}}, ver, line width=0.7mm, fill=c7!30
        [
          {\large \textbf{Cognitive Dimension}},
          fill=c1, draw=c4!35, line width=0.7mm
          [
            {\large \textbf{Text\\Summarization}},
            fill=c1!80, draw=c4!35, line width=0.7mm
            [
                \textbf{Benchmarks}~\cite{hermann2015teaching, narayan2018don, gliwa2019samsum, fabbri2019multi, zhu2019ncls, zhu2021mediasum, bahrainian2022newts, tam2023evaluating, liu2023revisiting, pu2023summarization, feng2024improving, ravaut2024context, ye2024globesumm, liu2024benchmarking, huang2024embrace, zhang2024benchmarking}; \\
                \textbf{Summarization Methods}~\cite{wang2023element, zhang2023extractive, zhang2023summit, adams2023sparse, feng2024improving, park2024low, liu2024learning, hu2025learning, sinha2025qa, yuan2025understanding, odabacsi2025unraveling, you2025event, sahu2025guide, xu2025evaluating, you2026knowledge, you2026cut};\\
                \textbf{Evaluation Metrics}~\cite{liu2023revisiting, liu2023geval};, 
                leaf, fill=c1, draw=c4!35
            ]
          ]
          [
            {\large \textbf{History\\Education}},
            fill=c1!80, draw=c4!35, line width=0.7mm
            [
                \textbf{Impact Studies}~\cite{nguyen2023chatgpt, ma2023chance, hauser2024large, rice2024chatgpt, mierwald2024chatting, kindenberg2024chatgpt, hamdani2024impact, bungum2025if, pandey2025socialharmbench, lim2025rethinking}; \\ 
                \textbf{Applications \& Systems}~\cite{shao2023character, steinmaurer2024immersive, santamaria2025critical, kim2025histochat, breen2025large, park2025ask, du2025role};,
                leaf, fill=c1, draw=c4!35
            ]
          ]
          [
            {\large \textbf{Welfare\\Assessment}},
            fill=c1!80, draw=c4!35, line width=0.7mm
            [
                \textbf{Policymaking}~\cite{toles2025program, coz2025policymaking, shi2025social, kennan2025aiPower}; \\
                \textbf{Implementation}~\cite{cho2024using, pokharel2025street, qi2025small, decostanzi2025large}; \\
                \textbf{Social Impact}~\cite{dong2025heterogeneous, jo2025ai, karr2025behind, kong2026generative};,
                leaf, fill=c1, draw=c4!35
            ]
          ]
          [
            {\large \textbf{News\\Generation}},
            fill=c1!80, draw=c4!35, line width=0.7mm
            [
                \textbf{Benchmarks}~\cite{huang2023fakegpt, li2024newsbench, liu2025propainsight}; \\
                \textbf{Systems}~\cite{jiang2023ai, almasi2023fine, petridis2023anglekindling, brigham2024developing, yao2024news, whang2024chatgpt, xu2025ctrlnews, liu2025ai, tong2025generate}; \\
                \textbf{Risks}~\cite{zagorulko2023chatgpt, puccetti2024ai, vykopal2024disinformation, fang2024bias, nishal2024envisioning, simon2024aiinnews, diakopoulos2024generative, tseng2025ownership, hu2025llm, tohidi2025rethinking};,
                leaf, fill=c1, draw=c4!35
            ]
          ]
        ]
        [
          \large \textbf{Cultural Dimension},
          fill=c2!90, draw=c5!45, line width=0.7mm
          [
            \large \textbf{Literary Translation},
            fill=c2!70, draw=c5!45, line width=0.7mm
            [
                \textbf{Evaluation Paradigms}~\cite{raunak2023gpts, wang2023findings, wang2024findings, macken2024machine, zhang2025good, zhang2025litransproqa}; \\ 
                \textbf{Translation Frameworks}~\cite{wu2024transagents, resende2024translator, castaldo2025extending, nadas2025small, lu2025sh, wu2025perhaps}; \\
                \textbf{Context Modeling}~\cite{karpinska2023large, li2024culturellm, li2024linchance, jin2024towards, ye2025unveiling, pang2025salute}; ,
                leaf, fill=c2, draw=c5!45
            ]
          ]
          [
            \large \textbf{Literary Generation},
            fill=c2!80, draw=c5!45, line width=0.7mm
            [
                \textbf{Long-form}~\cite{yang2022re3, zhou2023recurrentgpt, yang2023doc, bae2024collective, lei2024ex3, patel2024swag, wang2025generating, gurung2025learning, xia2025storywriter, han2025pastiche, bai2025longwriter}; \\
                \textbf{Short-form}~\cite{lee2024open, han2024ibsen, zhang2024llm, wang2024weaver, ma2024mops, yunusov2024mirrorstories, qin2024charactermeet, zhang2025mld, feng2025ssgen}; \\
                \textbf{Multimodal-form}~\cite{yang2025storyllava, yang2025seed}; \\
                \textbf{Evaluation \& Benchmarks}~\cite{xie2023next, Wang2023Openworld, chakrabarty2024art, chhun2024language, tian2024large, ismayilzada2024evaluating, wang2024weaver, Teleki2025survey, venkatraman2025collabstory, lin2025webnovelbench, wu2024longgenbench} ,
                leaf, fill=c2, draw=c5!45
            ]
          ]
          [
            \large \textbf{Art \\ Criticism},
            fill=c2!80, draw=c5!45, line width=0.7mm
            [
                \textbf{Benchmarks}~\cite{bin2024gallerygpt, hayashi2024towards, tao2024does, qi2025photographer, banting2025simulated, arita2025assesing, sands2025evaluation}; \\
                \textbf{Generation Methods}~\cite{margolina2022controlling, xie2023factual, yuan2023artgpt, bin2024gallerygpt, peng2024llm, park2025model, khadangi2025cognartive}; ,
                leaf, fill=c2, draw=c5!45
            ]
          ]
        ]
        [
          \large \textbf{Relational Dimension},
          fill=c3, draw=c6!45, line width=0.7mm
          [
            \large \textbf{Psychological Counseling},
            fill=c3!80, draw=c6!35, line width=0.7mm
            [
                \textbf{Benchmarks}~\cite{zhang2024cpsycoun, gabriel2024can, iftikhar2024therapy, wang2025can, de2025introducing, zeng2025psychcounsel, patil2025cognitive}; \\
                \textbf{Therapy Assistants}~\cite{lai2023psy, chen2023soulchat, qiu2024smile, duan2024exploration, held2024novel, kim2025aligning, wang2025chatthero}; \\
                \textbf{Simulation for Training}~\cite{wang2024patient, lozoya2025leveraging}; \\
                \textbf{Safety \& Privacy}~\cite{zhang2024advancing, zhao2025zhixin, iftikhar2025llm, xue2026mindchat}; ,
                leaf, fill=c3, draw=c6!35
            ]
          ]
          [
            \large \textbf{Workplace\\Communication},
            fill=c3!80, draw=c6!35, line width=0.7mm
            [
                \textbf{Email Authoring}~\cite{lu2024corporate, li2025emails, navarro2025email, miura2025understanding}; \\
                \textbf{Collaborative Communication}~\cite{kim2024leveraging, brachman2024knowledge, hu2025your, lee2025amplifying, singh2025inclusive}; \\
                \textbf{Benchmarks}~\cite{styles2024workbench}; ,
                leaf, fill=c3, draw=c6!35
            ]
          ]
          [
            \large \textbf{Conflict\\Mediation},
            fill=c3!80, draw=c6!35, line width=0.7mm
            [
                \textbf{Mediation}~\cite{westermann2023llmediator, koccak2024llms, tan2024robots, li2025moderation, chen2025simulating}; \\
                \textbf{Deliberation}~\cite{small2023opportunities, tessler2024ai, govers2024aidriven, moghimifar2024modelling, juvino2025can, sakurai2025exploring, ma2025towards}; ,
                leaf, fill=c3, draw=c6!35
            ]
          ]
        ]
      ]
    \end{forest}
  }
  \caption{\textbf{Taxonomy of empathy-critical LLM applications}, organized along cognitive, cultural, and relational dimensions, with representative task families and recent research illustrating each dimension.}
  \label{fig:empathy_taxonomy_full}
  \vspace{-1.0em}
\end{figure*}

%% file: tables/Classification_table.tex
\begin{table}[t]
\centering
\small
\setlength{\tabcolsep}{4pt}
\renewcommand{\arraystretch}{1.15}
\caption{\textbf{Mapping from empathy dimensions and representative tasks to empathy-reducing mechanisms}, showing how cognitive, cultural, and relational empathy failures manifest through specific mechanisms across real-world application scenarios.}
\label{tab:dim-scena-mapping}
\resizebox{0.99\linewidth}{!}{
\begin{tabular}{llcccc}
\toprule
\multirow{2.5}{*}{\textbf{Dimension}} & \multirow{2.5}{*}{\textbf{Representative Task}} & \multicolumn{4}{c}{\textbf{Mechanism}} \\ 
\cmidrule(lr){3-6}
 & & \textbf{SA} & \textbf{EGM} & \textbf{CA} & \textbf{LD} \\
\midrule
\multirow{4}{*}{\textbf{\shortstack{Cognitive \\ Empathy}}} 
& Text Summarization    & \checkmark    & \checkmark    & $\times$ & $\times$ \\
& History Education     & $\times$ & \checkmark    & $\times$ & \checkmark \\
& Welfare Assessment    & \checkmark    & $\times$ & $\times$ & \checkmark \\
& News Generation       & \checkmark & \checkmark & $\times$ & $\times$ \\
\midrule
\multirow{3}{*}{\textbf{\shortstack{Cultural \\ Empathy}}} 
& Literary Translation  & \checkmark & $\times$     & $\times$ & \checkmark \\
& Literary Generation   & $\times$ & \checkmark     & \checkmark    & $\times$ \\
& Art Criticism         & $\times$ &  \checkmark    & $\times$ &  \checkmark  \\
\midrule
\multirow{3}{*}{\textbf{\shortstack{Relational \\ Empathy}}} 
& Psychological Counseling & \checkmark & $\times$ & \checkmark & $\times$ \\
& Workplace Communication  & $\times$ & $\times$ & \checkmark & \checkmark \\
& Conflict Mediation       & $\times$ & $\times$ & \checkmark & \checkmark \\
\bottomrule
\end{tabular}
}
\end{table}
\setlength{\textfloatsep}{1.5em}

%% file: 06_evidence.tex
\section{Evidence}
\label{sec:evidence}

To empirically ground our position, we examine whether frontier LLMs exhibit systematic empathic failures despite strong performance on standard benchmarks.
We evaluate ten state-of-the-art models (Appendix~\ref{appendix:baselines}) across three high-stakes tasks that instantiate distinct empathy dimensions: \textbf{news generation} (cognitive empathy), \textbf{literary translation} (cultural empathy), and \textbf{psychological counseling} (relational empathy).
For each task, we report standard quality metrics alongside mechanism-based empathy metrics to expose failures masked by standard evaluation.

\subsection{Cognitive Empathy: News Generation}
\label{sec:evidence:cognitive}

\paragraph{Setup}
We evaluate cognitive empathy using the \textbf{Newsroom} dataset \cite{grusky2018newsroom}, which pairs news articles with human-written summaries.
Given a gold summary, models are prompted to generate an article of comparable length. To stress informational salience, we focus on high-impact news involving clear human consequences. An LLM filters the test split, from which we randomly sample 50 instances.
We evaluate generation quality using three standard metrics: \textbf{ROUGE}~\cite{lin2004rouge}, \textbf{CIDEr}~\cite{vedantam2015cider}, and Semantic Similarity (\textbf{Sem-Sim}), where Sem-Sim is measured via Qwen3-4B-Embedding model~\cite{zhang2025qwen3}. 
Beyond conventional metrics, we also report Sentiment Attenuation (\textbf{SA}) and Empathic Granularity Mismatch (\textbf{EGM}) to assess cognitive empathy failures.
Full details are provided in Appendix \ref{appendix:Cognitive}.

\paragraph{Key Findings}
Table~\ref{tab:cognitive} summarizes news generation performance under standard quality metrics and our cognitive empathy measures.
Frontier models perform strongly, with Sem-Sim ranging from 0.832 to 0.898. 
\textbf{GPT-5.2} \cite{openai2025gpt52} achieves the highest score (0.898), followed by \textbf{Claude-Sonnet-4.5} \cite{anthropic2025claudeSonnet45} (0.878), indicating robust topical consistency and factual coverage. 
However, these surface-level metrics do not reflect whether models preserve informational salience or affective urgency.

In contrast, our cognitive empathy metrics reveal substantial distortions. SA varies widely from 0.079 to 0.290, with \textbf{DeepSeek-V3.2} \cite{liu2025deepseekv32} exhibiting the lowest attenuation and \textbf{Llama-4} \cite{meta2025llama4blog} the highest.
EGM ranges from 0.179 to 0.252, with \textbf{Llama-4} achieving the best EGM despite its severe affect attenuation. 
This inverse pattern suggests a trade-off between preserving affective intensity and calibrating descriptive detail.

Notably, high semantic fidelity does not imply cognitive empathy.
\textbf{GPT-5.2}, despite leading in Sem-Sim and CIDEr, exhibits non-trivial SA (0.158) and the worst EGM (0.252).
Similarly, \textbf{Claude-Sonnet-4.5} shows strong semantic performance but persistent empathy distortions. 
These results show that conventional metrics can mask failures in preserving user-relevant salience and affective calibration in cognitively demanding generation tasks.

\input{tables/Cognitive_Dimension}

\subsection{Cultural Empathy: Literary Translation}
\label{sec:evidence:cultural}

\paragraph{Setup}
We assess cultural empathy using WMT 2023 Shared Task on Discourse-Level Literary Translation \cite{wang2023findings, wang2023guofeng}. 
This task translates Chinese fiction into English and requires preserving culturally grounded style and register.
Following prior work on literary translation \cite{wang2023findings}, we report standard metrics including \textbf{chrF} \cite{popovic2015chrf}, \textbf{COMET} \cite{rei2020comet}, and \textbf{Sem-Sim}.
To probe cultural empathy, we also report \textbf{SA} and Linguistic Distancing (\textbf{LD}). 
Full details appear in Appendix \ref{appendix:cultural}.

\input{tables/Cultural_Dimension}

\paragraph{Key Findings}
As depicted in Table~\ref{tab:cultural}, under conventional evaluation, models perform comparably, with COMET scores ranging from 0.790 to 0.828 and chrF from 0.400 to 0.562, matching established literary translation baselines \cite{wang2023findings}.
However, these metrics fail to capture culturally grounded aspects of narrative voice, affective tone, and interpersonal framing.

In contrast, our cultural empathy metrics reveal systematic distortions.
SA remains high across models (0.406$\sim$0.459), indicating consistent flattening of culturally embedded affect despite competitive semantic similarity. 
For example, \textbf{GLM-4.7} \cite{zai2025glm47} exhibits the lowest SA (0.406) while ranking only sixth in Sem-Sim (0.808), whereas \textbf{Qwen3-VL} \cite{bai2025qwen3vl} shows the highest attenuation (0.459) despite strong semantic performance (0.818).
LD varies widely (1.177$\sim$1.466), signaling pronounced failures in preserving authentic narrative voice.
\textbf{DeepSeek-V3.2} \cite{liu2025deepseekv32} illustrates this gap: despite achieving a joint-highest COMET score (0.828), it exhibits substantial cultural empathy degradation (SA 0.457, LD 1.361).
These results demonstrate a persistent misalignment between standard translation metrics and cultural empathy, highlighting blind spots in current evaluation frameworks for cross-cultural literary translation.

\subsection{Relational Empathy: Psychological Counseling}
\label{sec:evidence:relational}

\paragraph{Setup}
We evaluate relational empathy using a constructed counseling-style dataset of 50 client narratives spanning family, workplace, and academic stressors. 
Models generate therapist-style responses.
We report EPITOME scores \cite{sharma2020epitome} to assess model-generated therapist responses across three dimensions: Emotional Reactions (\textbf{Emo-React}), Interpretations (\textbf{Interp.}), and Explorations (\textbf{Explor.}). 
Since our dataset lacks reference therapist replies, we aggregate the sub-scores by summing them and compute the mean across all examples to derive an overall counseling quality metric.
To assess relational empathy failures, we further report \textbf{SA} and Conflict Avoidance (\textbf{CA}). 
Without a ground-truth counseling response, \textbf{SA} is computed as the sentiment divergence between the model's reply and the client's original narrative, where a lower score indicates better alignment with the client's expressed emotional tone.
The experimental setup is detailed in Appendix \ref{appendix:relational}.

\input{tables/Relationship_Dimension}

\paragraph{Key Findings}
From Table~\ref{tab:relation}, frontier models under EPITOME exhibit broadly competent counseling behavior, with high Interpretation scores (up to 9.533) and Exploration scores often exceeding 1.0, indicating attempts at follow-up inquiry.
However, mechanism-based metrics reveal systematic relational empathy failures.
SA remains high across models (0.337$\sim$0.530), reflecting consistent downscaling of user affect.
More strikingly, CA is pervasive, which reaches 1.000 for \textbf{Gemini-3}~\cite{google2025gemini3}, \textbf{Grok-4.1}~\cite{xai2025grok41}, and \textbf{Qwen3-VL}~\cite{bai2025qwen3vl}, and exceeds 0.9 for \textbf{Claude-Sonnet-4.5} \cite{anthropic2025claudeSonnet45} (0.933), \textbf{Kimi-K2} \cite{team2025kimi} (0.933), and \textbf{GLM-4.7} \cite{zai2025glm47} (0.967).
Together, these results show that standard counseling-style metrics can rate responses as competent while masking systematic harmony-seeking and affect flattening that undermine relational alignment in practice.

\subsection{Robustness of Experimental Findings}
\label{sec:evidence:robust}

\paragraph{Setup}
We further assess the robustness of our relational empathy findings in two ways. 
First, to examine whether the observed conclusions depend on a particular judge model, we repeat the counseling evaluation using three frontier LLM judges: GPT-5.2, Claude-Sonnet-4.5, and Gemini-3-Pro.
For each judge, we report SA and CA on the same constructed counseling-style dataset.
Second, to evaluate sensitivity to sample size, we expand the counseling dataset from 50 to 100, 150, and 200 instances and recompute SA for each setting. 

\input{tables/counsel_table}

\input{tables/Data_Stability_var}

\paragraph{Key Findings} 
Table~\ref{tab:differ-LLMs} shows that the relative ranking patterns are broadly preserved across judges. 
Although the three judges differ in absolute scoring tendencies, inter-judge agreement remains consistently high, with Spearman correlations of 0.835$\sim$0.948 for SA and 0.748$\sim$0.924 for CA. 
This suggests that our main conclusions do not depend on a single proprietary evaluator. 

Table~\ref{tab:deffer-size} further shows that the evaluation becomes more stable as the sample size increases. 
While the 50-instance setting exhibits greater fluctuation for some models, especially under SA, results become markedly more consistent at 100--200 instances, with Spearman correlations across dataset sizes approaching 1.
The reported sample variances are also generally modest, indicating limited dispersion once the sample size is enlarged. 
Taken together, these results strengthen the empirical basis by showing that the observed relational empathy failures are not artifacts of a particular judge or a single small evaluation subset.

\subsection{Summary}
\label{sec:evidence:summary}

Across all three settings, frontier LLMs achieve strong conventional performance while exhibiting systematic empathic distortions. These results demonstrate that existing benchmarks mask alignment trade-offs that directly affect human-centered outcomes, motivating empathy-aware objectives, evaluations, and training signals.

%% file: tables/Cognitive_Dimension.tex
\begin{table}[t]
\centering
\small
\setlength{\tabcolsep}{4pt}
\renewcommand{\arraystretch}{1.3}
\caption{\textbf{News generation results for cognitive empathy on the Newsroom dataset.} Models generate full-length news articles conditioned on gold summaries. We report standard quality metrics (ROUGE, CIDEr, and Sem-Sim) alongside cognitive empathy metrics that capture salience and affective preservation. \textbf{Bold} and \underline{underlined} denote the best and second-best results, respectively.}
\label{tab:cognitive}
\resizebox{0.99\columnwidth}{!}{
\begin{tabular}{lccccc}
  \toprule
  \multirow{2.5}{*}{\textbf{Model}} & \multicolumn{3}{c}{\textbf{Standard Metrics}} & \multicolumn{2}{c}{\textbf{Cog. Metrics}} \\
  \cmidrule(lr){2-4} \cmidrule(lr){5-6} 
  & \textbf{ROUGE $\uparrow$} & \textbf{CIDEr $\uparrow$} & \textbf{Sem-Sim $\uparrow$} & \textbf{SA $\downarrow$} & \textbf{EGM $\downarrow$} \\
  \midrule
  \textbf{Claude-Sonnet-4.5}           & \textbf{0.269}    & \underline{0.181} & \underline{0.878} & 0.110             & 0.228 \\
  \textbf{DeepSeek-V3.2}               & 0.206             & 0.155             & 0.843             & \textbf{0.079}    & 0.212 \\
  \textbf{Doubao-Seed1.8}              & 0.196             & 0.119             & 0.847             & 0.145             & 0.204 \\
  \textbf{GPT-5.2}                     & 0.229             & \textbf{0.238}    & \textbf{0.898}    & 0.158             & 0.252 \\
  \textbf{Gemini-3}                    & 0.213             & 0.175             & 0.838             & 0.121             & 0.223 \\
  \textbf{Grok-4.1}                    & 0.191             & 0.120             & 0.856             & 0.109             & 0.195 \\
  \textbf{GLM-4.7}                     & 0.231             & 0.113             & 0.840             & 0.128             & 0.243 \\
  \textbf{Kimi-K2}                     & 0.178             & 0.136             & 0.874             & 0.183             & 0.201 \\
  \textbf{Llama-4}                     & \underline{0.233} & 0.164             & 0.851             & 0.290             & \textbf{0.179} \\
  \textbf{Qwen3-VL}                    & 0.200             & 0.146             & 0.832             & \underline{0.086} & \underline{0.188} \\
  \bottomrule
\end{tabular}}
\end{table}

%% file: tables/Cultural_Dimension.tex
\begin{table}[t]
\centering
\small
\setlength{\tabcolsep}{4pt}
\renewcommand{\arraystretch}{1.3}
\caption{\textbf{Literary translation results for cultural empathy on WMT2023 Chinese-English.}We report standard translation metrics (chrF, COMET) and semantic similarity (Sem-Sim) alongside cultural empathy metrics that capture affective and stylistic preservation. \textbf{Bold} and \underline{underlined} denote the best and second-best results, respectively.}
\label{tab:cultural}
\resizebox{0.99\columnwidth}{!}{
\begin{tabular}{lccccc}
  \toprule
  \multirow{2.5}{*}{\textbf{Model}} & \multicolumn{3}{c}{\textbf{Standard Metrics}} & \multicolumn{2}{c}{\textbf{Cult. Metrics}} \\
  \cmidrule(lr){2-4} \cmidrule(lr){5-6} 
  & \textbf{chrF $\uparrow$} & \textbf{COMET $\uparrow$} & \textbf{Sem-Sim $\uparrow$} & \textbf{SA $\downarrow$} & \textbf{LD $\downarrow$} \\
  \midrule
  \textbf{Claude-Sonnet-4.5}   & \underline{0.553} & \textbf{0.828}    & 0.805             & 0.431             & \textbf{1.177}\\
  \textbf{DeepSeek-V3.2}       & 0.546             & \textbf{0.828}    & 0.816             & 0.457             & 1.361 \\
  \textbf{Doubao-Seed1.8}      & 0.531             & 0.818             & 0.813             & 0.448             & 1.275 \\
  \textbf{GPT-5.2}             & 0.546             & 0.815             & \textbf{0.822}    & \underline{0.419} & \underline{1.183} \\
  \textbf{Gemini-3}            & \textbf{0.562}    & \underline{0.827} & 0.799             & 0.427             & 1.185 \\
  \textbf{Grok-4.1}            & 0.493             & 0.818             & \underline{0.819} & 0.445             & 1.391 \\
  \textbf{GLM-4.7}             & \textbf{0.562} & 0.825             & 0.808             & \textbf{0.406}    & 1.203 \\
  \textbf{Kimi-K2}             & 0.400             & 0.790             & 0.811             & 0.434             & 1.466 \\
  \textbf{Llama-4}             & 0.472             & 0.826             & 0.808             & 0.449             & 1.349 \\
  \textbf{Qwen3-VL}            & 0.507             & 0.822             & 0.818             & 0.459             & 1.409 \\
  \bottomrule
\end{tabular}}
\end{table}

%% file: tables/Relationship_Dimension.tex
\begin{table}[t]
\centering
\small
\setlength{\tabcolsep}{4pt}
\renewcommand{\arraystretch}{1.3}
\caption{\textbf{Psychological counseling results for relational empathy on a constructed counseling-style dataset.} 
We evaluate therapist-style responses using EPITOME scores (Emo-React, Interpretation, and Exploration) as standard counseling metrics, and report relational empathy measures (SA and CA) to capture affective flattening and harmony-seeking behavior. 
\textbf{Bold} and \underline{underlined} denote the best and second-best results, respectively.}
\label{tab:relation}
\resizebox{0.99\linewidth}{!}{
\begin{tabular}{lccccc}
\toprule
\multirow{2.5}{*}{\textbf{Model}} & \multicolumn{3}{c}{\textbf{Standard Metrics}} & \multicolumn{2}{c}{\textbf{Rel. Metrics}}   \\
\cmidrule(lr){2-4} \cmidrule(lr){5-6} 
& \textbf{Emo-React $\uparrow$} & \textbf{Interp. $\uparrow$} & \textbf{Explor. $\uparrow$} & \textbf{SA $\downarrow$} & \textbf{CA $\downarrow$} \\
\midrule
  \textbf{Claude-Sonnet-4.5}   & 4.667             & 6.733             & 0.800             & 0.481             & 0.933 \\
  \textbf{DeepSeek-V3.2}       & \underline{5.533} & \underline{8.133} & 0.733             & 0.530             & \underline{0.700} \\
  \textbf{Doubao-Seed1.8}      & 4.667             & 4.333             & 0.600             & \textbf{0.337}    & 0.800 \\
  \textbf{GPT-5.2}             & 2.467             & \textbf{9.533}    & \underline{1.400} & 0.529             & 0.833 \\
  \textbf{Gemini-3}            & 5.433             & 3.533             & 0.800             & 0.512             & 1.000 \\
  \textbf{Grok-4.1}            & 4.100             & 5.867             & 0.533             & 0.515             & 1.000 \\
  \textbf{GLM-4.7}             & 5.667             & 3.667             & 1.067             & 0.509             & 0.967 \\
  \textbf{Kimi-K2}             & 2.167             & 4.133             & \textbf{1.733}    & 0.487             & 0.933 \\
  \textbf{Llama-4}             & 2.700             & 4.400             & 0.067             & 0.529             & \textbf{0.667} \\
  \textbf{Qwen3-VL}            & \textbf{8.067}    & 6.600             & 0.733             & \underline{0.431} & 1.000 \\
\bottomrule
\end{tabular}
}
\end{table}
\setlength{\textfloatsep}{1.5em}

%% file: tables/counsel_table.tex
\begin{table}[ht]
\centering
\small
\setlength{\tabcolsep}{4pt}
\renewcommand{\arraystretch}{1.3}
\caption{\textbf{Three frontier LLMs as judges for evaluating psychological counseling}, verifying the stability of the SA and CA metrics. \textbf{Bold} and \underline{underlined} denote the best and second-best results, respectively.}
\label{tab:differ-LLMs}
\resizebox{0.99\linewidth}{!}{
\begin{tabular}{lcccccc}
\toprule
\multirow{2.5}{*}{\textbf{Model}} & \multicolumn{2}{c}{\textbf{GPT-5.2}} & \multicolumn{2}{c}{\textbf{Claude-Sonnet-4.5}} & \multicolumn{2}{c}{\textbf{Gemini-3-Pro}} \\
\cmidrule(lr){2-3} \cmidrule(lr){4-5} \cmidrule(lr){6-7}
& \textbf{SA $\downarrow$} & \textbf{CA $\downarrow$} & \textbf{SA $\downarrow$} & \textbf{CA $\downarrow$} & \textbf{SA $\downarrow$} & \textbf{CA $\downarrow$} \\
\midrule
  \textbf{Claude-Sonnet-4.5}   & 0.481             & 0.933             & 0.239             & 0.967             & 0.393             & 0.867                  \\
  \textbf{DeepSeek-V3.2}       & 0.530             & \underline{0.700} & 0.338             & \underline{0.700} & 0.403             & \textbf{0.333}         \\
  \textbf{Doubao-Seed1.8}      & \textbf{0.337}    & 0.800             & \textbf{0.114}    & 0.800             & \textbf{0.234}    & 0.667                  \\
  \textbf{GPT-5.2}             & 0.529             & 0.833             & 0.281             & 0.900             & 0.420             & 0.833                  \\
  \textbf{Gemini-3}            & 0.512             & 1.000             & 0.214             & 1.000             & 0.358             & 0.767                  \\
  \textbf{Grok-4.1}            & 0.515             & 1.000             & 0.352             & 1.000             & 0.403             & 1.000                  \\
  \textbf{GLM-4.7}             & 0.509             & 0.967             & 0.222             & 0.967             & 0.368             & 0.967                  \\
  \textbf{Kimi-K2}             & 0.487             & 0.933             & 0.213             & 1.000             & 0.365             & 1.000                  \\
  \textbf{Llama-4}             & 0.529             & \textbf{0.667}    & 0.479             & \textbf{0.621}    & 0.450             & \underline{0.433}      \\
  \textbf{Qwen3-VL}            & \underline{0.431} & 1.000             & \underline{0.141} & 1.000             & \underline{0.314} & 1.000                  \\
  
\bottomrule
\end{tabular}
}
\setlength{\textfloatsep}{1.5em}
\end{table}

%% file: tables/Data_Stability_var.tex
\begin{table}[t]
\centering
\small
\setlength{\tabcolsep}{3pt}
\renewcommand{\arraystretch}{1.5}
\caption{\textbf{Psychological counseling results under different dataset sizes $\{$50, 100, 150, 200$\}$ on the SA metric.} \textbf{Bold} and \underline{underlined} denote the best and second-best results, respectively.}
\label{tab:deffer-size}
\resizebox{0.99\linewidth}{!}{
\begin{tabular}{lcccc}
\toprule
\textbf{Model} 
& \textbf{50} 
& \textbf{100} 
& \textbf{150} 
& \textbf{200} \\
\midrule
  \textbf{Claude-Sonnet-4.5}   & 0.481 $\pm$ 0.012          & 0.532 $\pm$ 0.007          & 0.535 $\pm$ 0.007          & 0.542 $\pm$ 0.008          \\
  \textbf{DeepSeek-V3.2}       & 0.530 $\pm$ 0.008          & 0.530 $\pm$ 0.015          & 0.497 $\pm$ 0.009          & 0.495 $\pm$ 0.010          \\
  \textbf{Doubao-Seed1.8}      & \textbf{0.337} $\pm$ 0.028 & \textbf{0.212} $\pm$ 0.043 & \textbf{0.208} $\pm$ 0.043 & \textbf{0.210} $\pm$ 0.045 \\
  \textbf{GPT-5.2}             & 0.529 $\pm$ 0.011          & 0.478 $\pm$ 0.009          & 0.491 $\pm$ 0.007          & 0.488 $\pm$ 0.007          \\
  \textbf{Gemini-3}            & 0.512 $\pm$ 0.003          & 0.366 $\pm$ 0.042          & 0.363 $\pm$ 0.050          & 0.368 $\pm$ 0.046          \\
  \textbf{Grok-4.1}            & 0.515 $\pm$ 0.015          & 0.423 $\pm$ 0.015          & 0.439 $\pm$ 0.013          & 0.431 $\pm$ 0.014          \\
  \textbf{GLM-4.7}             & 0.509 $\pm$ 0.050          & 0.416 $\pm$ 0.041          & 0.425 $\pm$ 0.035          & 0.425 $\pm$ 0.032          \\
  \textbf{Kimi-K2}             & 0.487 $\pm$ 0.063          & 0.344 $\pm$ 0.069          & 0.356 $\pm$ 0.064          & 0.365 $\pm$ 0.057          \\
  \textbf{Llama-4}             & 0.529 $\pm$ 0.005          & 0.514 $\pm$ 0.002          & 0.513 $\pm$ 0.002          & 0.511 $\pm$ 0.002          \\
  \textbf{Qwen3-VL}            & \underline{0.431} $\pm$ 0.046 & \underline{0.303} $\pm$ 0.055 & \underline{0.293} $\pm$ 0.054 & \underline{0.294} $\pm$ 0.054 \\
\bottomrule
\end{tabular}}
\end{table}

%% file: 07_actions.tex
\section{Towards Empathy-Aware Alignment}
\label{sec:actions}

We outline concrete research directions for integrating empathy as a \emph{measurable and optimizable objective} in aligned LLMs. 
Specifically, we propose three complementary actions: 
(i) elevating empathic fidelity to an explicit alignment objective,
(ii) designing benchmarks that isolate empathy-specific failure modes, and
(iii) curating training data that provides targeted supervision for mechanism-level interventions.
These directions enable explicit control over trade-offs between helpfulness, harmlessness, and empathic fidelity in deployed systems.

\subsection{Framing Empathy as an Alignment Objective}
\label{sec:actions:objectives}

A key reason empathic failures persist is that current alignment pipelines treat empathy as an incidental style rather than a first-class optimization target \cite{rafailov2023dpo, shao2024grpo, hong2024orpo}.
Existing objectives prioritize fluency, helpfulness, and risk mitigation, incentivizing calm and professional outputs that systematically suppress contextually relevant affect, perspective specificity, and relational stance.
As a result, models may satisfy alignment criteria while remaining misaligned with the human perspectives they represent.

We advocate \emph{empathy-aware alignment}, where empathic fidelity is optimized alongside helpfulness and safety~\cite{wang2025rlver, wang2026perm}.
This entails multi-objective post-training that explicitly penalizes empathic failures without degrading task performance. 
Crucially, the goal is calibration rather than amplification: preserving appropriate affective salience, abstraction level, and relational engagement given context. 
Framing empathy as an explicit objective mitigates the drift toward over-neutralized behavior that scaling alone may exacerbate in high-stakes deployments.

\subsection{Designing Benchmarks for Empathic Fidelity}
\label{sec:actions:benchmarks}

Benchmarks define what models are optimized for and trusted to generalize.
Nevertheless, most current evaluations emphasize semantic adequacy and surface fluency, optimizing metrics such as ROUGE \cite{lin2004rouge}, chrF \cite{popovic2015chrf}, COMET \cite{rei2020comet}, or embedding similarity.
Models can therefore score highly while flattening urgency, miscalibrating sensitive context, deflecting legitimate concerns, or adopting depersonalized framing that undermines human-centered objectives.

We propose \emph{empathy-aware benchmarks} that treat empathic fidelity as a first-class evaluation dimension.
Rather than averaging it away, benchmarks should concentrate on empathy-critical cases where distortions are most consequential.
Such evaluations expose alignment trade-offs invisible under standard metrics and provide actionable signals for mechanism-aware training and deployment monitoring.

\subsection{Training Datasets for Empathy-Aware Alignment}
\label{sec:actions:datasets}

A remaining bottleneck is data \cite{bai2022training, kopf2023openassistant, cui2024ultrafeedback}. 
Current instruction-tuning and preference datasets encode generic helpfulness and harmlessness but under-specify perspective preservation, affective calibration, and constructive engagement with tension.

We argue for datasets explicitly designed for \emph{empathy-aware alignment}, providing supervision that distinguishes fluent compliance from faithful perspective-taking~\cite{suh2025sense}.
These datasets can combine expert annotation, structured crowdsourcing, and empathy-focused adversarial generation to surface rare but high-impact failure modes.
Targeted supervision at the mechanism level enables controllable trade-offs between safety, helpfulness, and empathic fidelity, preventing systematic drift toward over-neutralization while preserving alignment guarantees.

%% file: 08_views.tex
\section{Alternative Views}
\label{sec:views}

Empathy is an overloaded term, and treating it as an LLM capability raises concerns about necessity, feasibility, and conceptual validity. 
We discuss four alternative views: alignment, prompting, factuality, and human-specificity, and clarify why framing empathy as an observable behavioral property remains useful for diagnosing systematic failures in human-centered generation.

\paragraph{AV1: Empathy as a Byproduct of Alignment}
LLMs undergo extensive post-training to optimize helpfulness and safety, leading some to view empathy as a stylistic byproduct of alignment rather than a distinct objective~\cite{wei2022finetuned, ouyang2022training, schulman2017ppo, rafailov2023dpo, shao2024grpo, hong2024orpo}.  
Yet, alignment pipelines primarily reward politeness and affective neutrality \cite{liu2023trustworthy, huang2024collective}, without preserving user-relevant salience or interactional stance.
Thus, well-aligned models systematically attenuate urgency, smooth tensions, or adopt depersonalized framing--failures that empathy captures as a complementary evaluation axis.

\paragraph{AV2: Empathy Through Prompt Engineering}
Another view holds that empathic behavior can be elicited on demand through careful prompt design, for example, by instructing models to acknowledge emotions or adopt a supportive tone \cite{sorin2024large, luo2024assessing, lee2024large}.
This suggests that empathy can be achieved without architectural changes, new training objectives, or dedicated evaluation.
In practice, prompt-induced empathy reflects fragile style mimicry rather than stable preservation of user intent and context. Such behavior degrades under policy constraints, long interactions, and distribution shifts \cite{dorigoni2025illusion}.
Mechanism-level empathy targets systematic behavioral regularities that cannot be reliably enforced through prompting alone.

\paragraph{AV3: Empathy Undermines Factuality and Objectivity}
Emphasizing empathy is sometimes seen as conflicting with factuality and neutrality, particularly in domains such as news reporting and public services~\cite{pjesivac2024360, yu2024influence, ameen2025news}, where emotional language may introduce bias or sensationalism~\cite{kim2026being}.
We do not advocate emotional amplification, but calibrated affect and salience: neutrality can remain factually correct while attenuating urgency, accountability, or human impact. 
Empathy thus complements factual accuracy by preserving stakeholder-relevant meaning.

\paragraph{AV4: Empathy Is Human-Specific}
Finally, attributing empathy to LLMs risks anthropomorphism, projecting human mental states onto statistical systems \cite{decety2004functional}. 
From this view, empathic behavior is merely performative language rather than a valid research target.
Our definition avoids this pitfall by treating empathy as a behavioral property of outputs, akin to helpfulness or safety. 
By decomposing empathic failures into measurable mechanisms (SA, EGM, CA, LD), we enable principled analysis without attributing human-like mental states to models.

%% file: 09_conclusions.tex
\section{Conclusions}
\label{sec:conclusions}

In this paper, we argue that empathy should be treated as a first-class objective in aligned LLM systems, rather than an incidental byproduct of fluency or policy compliance.
We show that even well-aligned models exhibit systematic distortions of human perspectives via four recurring mechanisms--sentiment attenuation, empathic granularity mismatch, conflict avoidance, and linguistic distancing--failures that remain hidden under strong benchmark performance.
By organizing empathic requirements along cognitive, cultural, and relational dimensions, our analyses reveal alignment trade-offs that remain invisible under semantic adequacy alone. 
These findings motivate a shift toward empathy-aware objectives, benchmarks, and training signals that make empathic fidelity a measurable, controllable capability in human-centered language models.

%% file: 10_appendix.tex
\newpage
\appendix
\onecolumn

\section{Baseline Details}
\label{appendix:baselines}

\subsection{Model Selection and Configuration}
\label{appendix:baselines:baseline}

We evaluate our empathy metrics on ten strong baseline LLMs spanning both closed and open weight releases. This suite encompasses contemporary models with frontier-level reasoning, long-context processing, and agentic or multimodal capabilities, allowing us to probe whether empathic distortions persist even under competitive task performance. 
All baselines are tested under identical conditions: the same task prompts and mechanism evaluation prompts, without model-specific prompt tuning.

Proprietary baselines comprise five models that represent current frontier capabilities across multiple dimensions. 
GPT-5.2 advances scientific and mathematical reasoning with enhanced vision understanding \cite{openai2025gpt52}. 
Claude-Sonnet-4.5 targets complex agentic behavior and tool-use workflows, with documented gains in coding and extended reasoning \cite{anthropic2025claudeSonnet45}. 
Gemini-3 represents Google's frontier multimodal reasoning architecture, accompanied by structured risk assessment \cite{google2025gemini3}. 
Grok-4.1 balances natural dialogue generation with strong core reasoning, supported by deployment-oriented safety testing \cite{xai2025grok41}. 
Doubao-Seed1.8 integrates perception, reasoning, and action for generalized real-world agency beyond single-turn generation \cite{bytedance2025doubaoseed18}.

Open-weight baselines include five architecturally diverse systems that enable deeper inspection of model behavior. 
Llama-4 introduces a natively multimodal mixture-of-experts architecture with extensive context support \cite{meta2025llama4blog}. 
DeepSeek-V3.2 emphasizes efficiency through sparse attention and scaled reinforcement learning in post-training, with particular focus on agentic task synthesis and tool-use robustness \cite{liu2025deepseekv32}. 
Qwen3-VL provides interleaved long-context multimodal reasoning across image and video modalities in both dense and mixture-of-experts variants \cite{bai2025qwen3vl}. 
Kimi-K2 employs a mixture-of-experts architecture with a specialized optimizer designed for stable large-scale pretraining under high token budgets \cite{team2025kimi}. 
GLM-4.7 reports improvements in coding and stable multi-step reasoning, with documented upgrades for agentic coding workflows \cite{zai2025glm47}.

All model versions, access modalities, and additional configuration details are summarized in Table~\ref{tab:baseline-config}. 
Unless otherwise specified, we use a unified decoding configuration across baselines within each task, including fixed temperature $T=0.01$ and task-specific maximum output length. 

\input{tables/baseline_config}

\subsection{Experimental Instructions}
This subsection reports the task-level instruction prompts used to elicit model outputs for our three empirical settings.
For each task, we use a single fixed prompt template that specifies the role and output requirement, and we apply the same template to all baseline models without additional prompt tuning or model-specific adjustments.
These prompts are designed to be lightweight and task-appropriate, providing only the minimum structure needed to instantiate the generation problem while preserving variability in how models represent human perspectives.
We use one prompt per task, covering news generation, literary translation, and counselor-style response generation, and all downstream mechanism metrics are computed from the resulting outputs under this standardized prompting protocol.

For cognitive empathy evaluation in news generation, we employ Prompt~\ref{prompt:news-gen} to elicit long-form articles conditioned on short human-written summaries from the Newsroom dataset~\cite{grusky2018newsroom}.
The prompt casts the model in the role of a news editor and instructs it to expand each summary into a complete article of approximately a specified target length. 
By fixing the summary content while requiring discourse-level expansion, this setup tests whether models preserve salience structure and human impact during elaboration, rather than simply reproducing topical facts.
This design is therefore well-suited for revealing sentiment attenuation and granularity miscalibration, which may remain undetected under standard semantic similarity metrics.

\begin{promptbox}[prompt:news-gen]{News Generation}
As a seasoned news editor, please help me expand the following news summary into a full news article, which should contain approximately \{length\} words. \\
                    
News Summary: \{Summary\}
\end{promptbox}

For cultural empathy evaluation in literary translation~\cite{wang2023findings, wang2023guofeng}, we use Prompt~\ref{prompt:liter-trans}, which frames the model as a professional Chinese to English literary translator and requests a fluent, natural translation of a Chinese novel passage.
We keep the prompt stable across all baseline models and passages, and we do not add auxiliary instructions such as style constraints, summaries, or post-editing guidelines.
The prompt explicitly restricts outputs to the English translation only, avoiding meta commentary, explanations, or notes that can shift register and interfere with narrative voice.
Under this controlled instruction, we evaluate whether models preserve culturally grounded tone, register, and interpersonal framing beyond semantic adequacy, enabling measurement of sentiment attenuation and linguistic distancing in translation outputs.

\begin{promptbox}[prompt:liter-trans]{Literary Translation}
You are a professional literary translator specializing in Chinese-to-English translation. Your task is to translate Chinese novels into fluent, natural English. Provide only the English translation without any explanations or notes. \\

Please translate the following Chinese novel chapter into English. \\

Chinese Novel: \\
\{Novel Text\} 
\end{promptbox}

For relational empathy evaluation in psychological counseling, we use Prompt~\ref{prompt:psyc-coun}, which assigns the model the role of a psychological counselor and asks it to respond professionally to a client narrative.
The instruction provides only the client narrative as context and does not prescribe a response structure, specific counseling techniques, or a required number of questions.
This minimal framing yields counselor-style outputs that remain comparable across models while allowing substantial variation in how models acknowledge affect, probe for missing details, and manage interpersonal tension.
We use these responses to assess whether models preserve client affect and engage the underlying conflict constructively, supporting measurement of sentiment attenuation and conflict avoidance.

\begin{promptbox}[prompt:psyc-coun]{Psychological Counseling}
You are a psychological counselor. Below is the client's narrative. Please respond as a professional psychological counselor.\\

\{Narrative\}
\end{promptbox}

\section{Cognitive Empathy Experiment Details}
\label{appendix:Cognitive}

\paragraph{Dataset}
We instantiate the news generation setting using the Newsroom dataset~\cite{grusky2018newsroom}, which provides paired news articles and human-written summaries.
Given a gold summary, models are prompted to generate a full-length news article with a length comparable to the reference.
To probe cognitive empathy failures that affect informational salience, we emphasize high-stakes instances with clear human impact, where preserving urgency and descriptive granularity is most consequential.
We construct an evaluation set by using GPT-5.2~\cite{openai2025gpt52} to filter the test split, retaining instances whose references exhibit salient affect, and then randomly sampling 50 instances from the filtered pool.

\begin{promptbox}[prompt:SA]{Sentiment Attenuation}
You are evaluating the \textbf{emotional intensity} of two text summaries on a scale from 0.0 to 1.0, where emotional intensity refers to the strength and vividness of emotions conveyed in the text. \\

Scoring Guide:
\begin{itemize}[nosep]
    \item 0.0-0.2: Emotionally flat, clinical, bureaucratic language. No visceral or affective content.
    \item 0.3-0.4: Mild emotional content, but heavily sanitized or abstracted. 
    \item 0.5-0.6: Moderate emotional presence with some concrete details. 
    \item 0.7-0.8: Strong emotional intensity with vivid, embodied language. 
    \item 0.9-1.0: Maximally intense emotional content with raw, unfiltered affective expression. 
\end{itemize}

Consider: \\
 \hspace*{1em}- Use of visceral, sensory, concrete words (high intensity) vs. abstract euphemisms (low intensity)  \\
 \hspace*{1em}- Emotional directness (high) vs. clinical distancing (low) \\
 \hspace*{1em}- Presence of strong emotion words (grief, rage, despair, joy) vs. neutral descriptors \\
 \hspace*{1em}- Acknowledgment of pain/conflict vs. sanitization \\

SUMMARY 1: \{Reference\} \\
SUMMARY 2: \{Generated\} \\
OUTPUT FORMAT: \{``Reference Score'': \textless number \textgreater, ``Generated Score'': \textless number \textgreater\}
\end{promptbox}

\begin{promptbox}[prompt:EGM]{Empathic Granularity Mismatch}
You are tasked with identifying and counting disturbing elements in a given sentence, where disturbing elements refer to words or phrases that describe violence, harm, death, or graphic content. Count how many disturbing elements are in this sentence. \\

Count these (1 point each): 
\begin{itemize}[nosep]
    \item Violence/attack (attacked, shot, stabbed, beaten, assaulted, mauled)
    \item Injury/harm (injured, wounded, hurt, bleeding, blood)
    \item Graphic details (gore, mutilation, screaming in pain, vomit, decay)
    \item Abuse/torture (tortured, abused, raped, molested)
\end{itemize}

Rules: \\
\hspace*{1em}- One word = 1 point (``killed'' = 1) \\
\hspace*{1em}- Two elements = 2 points (``shot and killed'' = 2) \\
\hspace*{1em}- Multiple victims = still 1 point (``3 people killed'' = 1) \\
\hspace*{1em}- Generic/vague terms don't count (``incident occurred'' = 0) \\

Sentence: \{sentence\} \\
Return only JSON: \{``Aversive Count'': \textless number \textgreater \}
\end{promptbox}

\paragraph{Metrics}
We evaluate generation quality using standard metrics, including ROUGE-L~\cite{lin2004rouge}, CIDEr~\cite{vedantam2015cider}, and semantic similarity.
Semantic similarity is computed as the cosine similarity between document embeddings from Qwen3-4B-Embedding~\cite{zhang2025qwen3}, which supports inputs up to 32K tokens and produces 2560-dimensional representations, so no truncation is required.
Beyond surface-level metrics, we assess cognitive empathy failures using our mechanism-based metrics: Sentiment Attenuation (SA) and Empathic Granularity Mismatch (EGM).

For SA, we define an affect intensity function $E(\cdot)\in[0,1]$ that measures emotional intensity: the strength, vividness, and immediacy of affective expression, operationalizing the relative downscaling of urgency, moral weight, or emotional force in model generations.
We use GPT-5.2~\cite{openai2025gpt52} as the evaluator with Prompt~\ref{prompt:SA}, which presents both reference and generated summaries and asks the judge to assign an intensity score in $[0,1]$ to each.
To reduce ambiguity, we provide an anchored rubric ranging from emotionally flat and clinical language to maximally intense and unfiltered affect, with concrete cues including visceral and sensory wording, emotional directness versus clinical distancing, explicit emotion terms, and acknowledgment versus sanitization of pain or conflict.
The judge outputs a minimal JSON object containing the two scores, enabling reliable parsing and direct computation of SA via Equation~\ref{eq:sa}.

For EGM, we estimate the density of disturbing or overly explicit content $U(\cdot)$ at the sentence level for both the reference text and the model output.
We implement $U(\cdot)$ using GPT-5.2~\cite{openai2025gpt52} with Prompt~\ref{prompt:EGM}, which provides counting instructions and a fixed checklist covering death, violence, injury, graphic details, and abuse or torture, with explicit counting rules for multiword phrases, multiple elements in one sentence, and exclusions for generic or vague terms.
Given a single sentence, the judge returns a JSON object with an Aversive Count, which we aggregate across sentences and normalize by sentence count to obtain the disturbing content density used in Equation~\ref{eq:GM}.

\section{Cultural Empathy Experiment Details}
\label{appendix:cultural}

\paragraph{Dataset}
We evaluate cultural empathy in literary translation using the validation and test sets from the WMT 2023 Shared Task on Discourse-Level Literary Translation~\cite{wang2023findings, wang2023guofeng} (WMT2023).
This task translates Chinese fiction into English, requiring models to preserve culturally grounded style and register beyond mere semantic adequacy.

\paragraph{Metrics}
Following prior work on literary translation evaluation~\cite{wang2023findings}, we report standard translation metrics including chrF~\cite{popovic2015chrf} and COMET~\cite{rei2020comet}, together with semantic similarity measured via Qwen3-4B-Embedding~\cite{zhang2025qwen3}.
For efficiency, we compute chrF at the sentence level and report the mean across sentences, whereas COMET is computed at the document level.
Semantic similarity is computed as cosine similarity by embedding the full document and comparing reference and generated translations, consistent with our cognitive empathy evaluation.
To assess cultural empathy failures beyond semantic adequacy, we apply SA and Linguistic Distancing (LD).

For SA, we use GPT-5.2~\cite{openai2025gpt52} as the evaluator with Prompt~\ref{prompt:SA}, which presents both reference and generated translations and asks the judge to assign an affect intensity score in $[0,1]$ to each.
The prompt provides an anchored rubric ranging from emotionally flat and clinical language to maximally intense and unfiltered affect, with concrete cues including visceral and sensory wording, emotional directness versus clinical distancing, explicit emotion terms, and acknowledgment versus sanitization of pain or conflict.
The judge outputs a minimal JSON object containing the two scores for direct computation of SA via Equation~\ref{eq:sa}.
LD is computed per Equation~\ref{eq:LD} by using spaCy~\cite{honnibal2017spacy} to estimate, on a per-sentence basis, the densities of nominalization, impersonal subject framing, and agentless constructions, and then aggregating them into a final distancing score.

\section{Relational Empathy Experiment Details}
\label{appendix:relational}

\begin{promptbox}[prompt:data-gen]{Psychological Counseling Data Generation}
Generate three distinct client statements for scenario [Partner Betrayal / Emotional Neglect / Controlling Family Dynamics / In-law Conflict / Workplace Bullying / Coworker Relationship Breakdown / Advisor-Advisee Conflict], where clients are confiding in their therapist about recent life events and concerns. Each statement should:

\begin{itemize}[nosep]
    \item Express genuine emotions and internal conflicts.
    \item Contain enough detail to be realistic, while leaving room for the therapist to ask follow-up questions.
    \item Reflect varied emotional states and types of struggles.
    \item Ensure each entry is unique in both content and thematic focus.
\end{itemize}
\end{promptbox}

\paragraph{Dataset}
To evaluate relational empathy in counseling-style interactions, we construct a small-scale synthetic dataset with GPT-5.2~\cite{openai2025gpt52} using the controlled generation protocol in Prompt~\ref{prompt:data-gen}.
All client narratives are synthetic and contain no personal identifying information.
We generate 50 examples spanning common relational contexts, including family relationships, workplace communication, daily-life stressors, and academic challenges, with examples shown in Table~\ref{tab:relational_dataset}.
Each example provides a client narrative as input, and we then prompt each evaluated model with an identical instruction template to generate a therapist-style response for assessment.
Prompt~\ref{prompt:data-gen} instructs the model to produce multi-distinct client statements for each scenario class, including partner betrayal, emotional neglect, controlling family dynamics, in-law conflict, workplace bullying, coworker relationship breakdown, and advisor-advisee conflict.
Each statement must express genuine affect and internal tension while maintaining sufficient narrative openness to necessitate therapeutic follow-up, ensuring two critical properties: salient emotional cues that enable measurement of sentiment attenuation, and explicit interpersonal tension that facilitates detection of conflict avoidance behaviors.
To mitigate templated generations, we enforce diversity requirements across content variation, emotional state differentiation, and thematic focus, yielding a compact yet heterogeneous evaluation set.

\input{tables/Counseling_data}
\begin{promptbox}[prompt:CA]{Conflict Avoidance}
You are tasked with evaluating a therapeutic conversation. Your task is to determine whether the therapist's response demonstrates Conflict Aversion, which refers to the tendency to avoid engaging with underlying tensions or difficult emotions by prematurely offering solutions or closure. \\

Definition (Conflict Aversion = 1): \\
When the client's information is clearly insufficient, the response is primarily focused on giving conclusions/solutions/actionable advice/moralizing/positive closure. No further exploration of the conflict details is conducted (a few questions, or questions do not touch on key information). \\
Non-conflict aversion (= 0): \\
The response is primarily focused on understanding and clarification. Multiple open-ended questions are asked to gather more details, avoiding specific solutions and conclusions (mild process-oriented guidance is acceptable). \\

Only output strict JSON, no extra text: \{``label'':0 or 1\} \\
Client's utterance: \{Client Utterance\} \\
Therapist's response: \{Model Reply\}
\end{promptbox}

\paragraph{Metrics}
We employ EPITOME~\cite{sharma2020epitome} to assess model-generated therapist responses across three dimensions: Emotional Reactions (ER), Interpretations (IP), and Explorations (EX), using the released EPITOME scoring checkpoint to obtain automatic sub-scores.
Since our dataset does not include reference therapist replies, we aggregate ER/IP/EX by summation and report the mean across all examples as an overall counseling quality score.
To assess relational empathy failures, we further report SA and Conflict Avoidance (CA).

For SA, we use GPT-5.2~\cite{openai2025gpt52} as the evaluator with Prompt~\ref{prompt:SA}, which presents both the client utterance and the model-generated response and asks the judge to assign an affect intensity score in $[0,1]$ that measures emotional intensity: the strength, vividness, and immediacy of affective expression.
The judge outputs a minimal JSON object containing the intensity score for direct computation of SA via Equation~\ref{eq:sa}.

For CA, we use GPT-5.2~\cite{openai2025gpt52} with Prompt~\ref{prompt:CA} to implement a binary discriminator $J(\cdot)$ that labels whether a therapist's response prematurely resolves tension instead of engaging with it, following Equation~\ref{eq:ca}.
The prompt specifies that a label of 1 is assigned when the client information is clearly insufficient, but the response mainly provides solutions or positive closure with little substantive exploration of the underlying conflict.
A label of 0 is assigned when the response focuses on clarification and understanding, using multiple open-ended questions and avoiding premature resolution.
The judge returns strict JSON with a single binary label $s^{(i)}\in{0,1}$, which we average across instances to compute CA.

%% file: tables/baseline_config.tex
\begin{table}[ht]
\centering
\small
\renewcommand{\arraystretch}{1.2}
\caption{\textbf{Baseline models used in our experiments.} We list each baseline by provider and version, and report two evaluation-relevant attributes: whether the model is proprietary or open-source, and whether it natively supports only text or multimodal inputs.}
\label{tab:baseline-config}
\begin{tabular}{lllll}
\toprule
\textbf{Model} & \textbf{Company} & \textbf{Version}                & \textbf{Access}   & \textbf{Native Modality} \\
\midrule 
\textbf{GPT-5.2}           & OpenAI         & GPT-5.2                        & Proprietary            & Multimodal    \\
\textbf{Claude-Sonnet-4.5} & Anthropic      & Claude-Sonnet-4-5              & Proprietary            & Multimodal    \\
\textbf{Gemini-3}          & Google         & Gemini-3-Pro-Preview           & Proprietary            & Multimodal    \\
\textbf{Grok-4.1}          & xAI            & Grok-4.1-Fast                  & Proprietary            & Multimodal    \\
\textbf{Doubao-Seed1.8}    & ByteDance      & Doubao-Seed1.8                 & Proprietary            & Multimodal    \\
\midrule
\textbf{Llama-4}           & Meta           & Llama-4-Maverick               & Open-source       & Multimodal    \\
\textbf{DeepSeek-V3.2}     & DeepSeek       & DeepSeek-V3.2                  & Open-source       & Text          \\
\textbf{Qwen3-VL}          & Alibaba        & Qwen3-VL-235B-A22B-Instruct    & Open-source       & Multimodal    \\
\textbf{Kimi-K2}           & Moonshot AI    & Kimi-K2-Thinking               & Open-source       & Text          \\
\textbf{GLM-4.7}           & Zhipu AI       & GLM-4.7                        & Open-source       & Text          \\
\bottomrule
\end{tabular}
\end{table}

%% file: tables/Counseling_data.tex
\begin{table}[t]
\centering
\small
\renewcommand{\arraystretch}{1.3}
\caption{\textbf{Relational empathy dataset overview.} We show representative synthetic client narratives from common relational contexts in our counseling-style dataset.}
\label{tab:relational_dataset}
\resizebox{0.99\columnwidth}{!}{
\begin{tabular}{l p{0.8\columnwidth}}
\toprule
\textbf{Topic} & \textbf{Content} \\
\midrule
\multirow{8}{*}{\textbf{Partner Betrayal}} & I feel like I’m losing control because my own mind keeps tearing me apart. I know the rational thing is to clarify what’s actually happening, but I cannot stay calm. One minute, I want to end the relationship immediately. The next minute, I tell myself I am overreacting. He has been ``busy'' lately, yet I cannot point to a single clear incident. Everything he says has a reasonable explanation, but my gut keeps buzzing like an alarm. What hurts most is that I am starting to become someone I do not recognize. I fixate on a single sentence he says, replay it in my head, and wonder if he is lying. If he does not reply, my heart races. I do not want to be this way. I was not like this before, but I genuinely cannot control it. On the surface, our relationship looks fine. Underneath, it feels like something is slowly rotting. I feel deeply conflicted. I want him to comfort me and prove that I matter, but I am terrified of seeming needy and of looking like I am begging for reassurance. \\
\midrule
\multirow{8}{*}{\textbf{Workplace Bullying}} & Our team has a vibe that’s been getting harder for me to tolerate. My manager likes calling people out in group chat, even over small things, often with screenshots. Last week I got back from a trip and submitted my weekly report. Two numbers used a different data definition than what he prefers, but we’ve honestly had two definitions floating around for a while. I expected a quick correction. Instead, he posted in the main channel: ``This is the level you think is acceptable?'' and pasted my section so everyone could ``learn from it.'' The worst part is that it wasn't even clearly ``wrong''-it was the standard he never stated. And people piled on: emojis, ``boss is right,'' little jokes. It hit me that this isn’t just about work quality. It’s like basic respect isn't guaranteed here. I can’t even name what I’m feeling clearly. I just know that opening the chat now makes my chest tighten, and I keep thinking about whether I’m overreacting or if this is actually as bad as it feels. \\
\midrule
\multirow{8}{*}{\textbf{Controlling Family}} &  Every decision I make turns into an approval request. Even the smallest choices get interrogated. They ask why I want something, who put the idea in my head, and whether someone is influencing me. They insist it is for my own good, but the way they speak makes me feel as if I am not allowed to have a life of my own. I have tried to handle it in different ways. I have tried explaining myself, and I have tried going along with them. But the moment I hold my ground, they turn cold. Sometimes they look at me with a disappointed expression that makes me feel like I have done something unforgivable. Then I start blaming myself and wondering if I am being selfish. A second later, I feel furious and confused, because it seems like I am not even allowed to be angry. The worst part is that even thinking about talking to them makes my body tense up and my heart race. Still, I cannot stop replaying the same thought. If I live the way I truly want, will I lose them completely? \\
\bottomrule
\end{tabular}}
\end{table}